# Fight sample degeneracy and impoverishment in particle filters: A review of intelligent approaches

**Tiancheng Li**
T. Li is with the School of Mechatronics, Northwestern Polytechnical University, Xi'an, 710072, China. (Corresponding author to provide e-mail: t.c.li@mail.nwpu.edu.cn; lit3@lsbu.ac.uk; Tel: +86(029) 8849 4701)
**Shudong Sun**
S. Sun is with the School of Mechatronics, Northwestern Polytechnical University, Xi'an, 710072, China (e-mail: sdsun@nwpu.edu.cn)
**Tariq Pervez Sattar**
T. P. Sattar is with the Center for Automated and Robotics NDT, London South Bank University, London, SE1 0AA, UK (e-mail: sattartp@lsbu.ac.uk)
**Juan Manuel Corchado**
Bioinformatic, Intelligent Systems and Educational Technology (BISITE) / Biomedical Research Institute of Salamanca (IBSAL), University of Salamanca, Plaza de la Merced S/N, 37008, Salamanca, Spain (e-mail: corchado@usal.es)

*Abstract*—during the last two decades there has been a growing interest in Particle Filtering (PF). However, PF suffers from two long-standing problems that are referred to as sample degeneracy and impoverishment. We are investigating methods that are particularly efficient at Particle Distribution Optimization (PDO) to fight sample degeneracy and impoverishment, with an emphasis on intelligence choices. These methods benefit from such methods as Markov Chain Monte Carlo methods, Mean-shift algorithms, artificial intelligence algorithms (e.g., Particle Swarm Optimization, Genetic Algorithm and Ant Colony Optimization), machine learning approaches (e.g., clustering, splitting and merging) and their hybrids, forming a coherent standpoint to enhance the particle filter. The working mechanism, interrelationship, pros and cons of these approaches are provided. In addition, Approaches that are effective for dealing with high-dimensionality are reviewed. While improving the filter performance in terms of accuracy, robustness and convergence, it is noted that advanced techniques employed in PF often causes additional computational requirement that will in turn sacrifice improvement obtained in real life filtering. This fact, hidden in pure simulations, deserves the attention of the users and designers of new filters.

## 1. Introduction

The Sequential Monte Carlo (SMC) approach allows inference of full posterior distributions via Bayesian filtering in general nonlinear state-space models where the noises of the model can be non-Gaussian. There has been great interest in applying the SMC approach to deal with a wide variety of nonlinear filtering problems. This method is normally called the Particle Filter(ing) (PF) [1], also referred to as Sequential imputations [2], the Monte Carlo filter [3], the Condensation filter [4], and the survival of fittest and the likelihood weighting algorithm [5]. To date, particle filters have been successfully applied in different areas including finance [6], parameter estimation [7, 19], geophysical systems [8], wireless communication [9], decision making [10, 21], tracking and defense [11, 23, 31], robotics [12] and some nontrivial applications [18, 19]. Additionally, a variety of strategies have been proposed to improve the performance of the particle filter in terms of accuracy, convergence, computational speed, etc. The staged survey of different years can be seen; examples include 2000[13], 2002 [14], 2007[15], 2009[16], 2010[17], etc. However, PF continues to suffer from two notorious problems: sample degeneracy and impoverishment, which is arguably a long-standing topic in the community. A variety of methods have been investigated to fight these two problems in order to combat the weakness of the particle filter.

This study does not purport to give either a comprehensive review of the development of general particle filters or its special applications. Both are covered in the aforementioned survey papers. Our aim is to investigate a group of emerging 'intelligent' ways employed within PF that have benefited from a variety of intelligent and heuristic algorithms. These variations of techniques, acting in different ways to optimize the spatial distribution of particles namely Particle Distribution Optimization (PDO), are particularly effective in alleviating sample degeneracy and impoverishment, forming a very systematic standpoint that is both mathematically sound and practically efficient to enhance PF. In addition, approaches that are effective in dealing with high-dimensional filtering, another obstacle for the SMC, are reviewed. This study aims to coordinate these developments into a unifying framework, unveiling their pros and cons and thereby directing further improvements of existing schemes. This survey is specifically expected to serve as the first comprehensive coverage of artificial intelligence and machine learning techniques applied in PF.

The basic background of PF is presented in section 2 with emphasis on its two fundamental difficulties: sample degeneracy and impoverishment. These two difficulties have motivated the development of a variety of PDO approaches, which are reviewed in the categories identified in section 3. Further discussions on the PDO framework including the computational efficiency and high dimensionality challenge are given in section 4. The conclusion is given in section 5.

## 2. Sample degeneracy and impoverishment

Before we proceed, we provide a brief review of PF and define the notation. The primary notations used are summarized in Table I.

Table I. Primary Notations

| | |
|---|---|
| $x_t$ | The state of interest at time $t$ |
| $X_t$ | $X_t \triangleq (x_0, x_1, \dots x_t)$, the history path of the state |
| $y_t$ | Observation at time $t$ |
| $Y_t$ | $Y_t \triangleq (y_0, y_1, \dots y_t)$, the history path of the observation |
| $g_t(\cdot)$ | the state transition equation at time $t$ |
| $h_t(\cdot)$ | the observation equation at time $t$ |
| $u_t$ | Noise affecting the system dynamic equation $g_t(\cdot)$, at time $t$ |
| $v_t$ | Noise affecting the observation equation $h_t(\cdot)$, at time $t$ |
| $x_t^{(i)}$ | The state of particle $i$, at time $t$ |
| $w_t^{(i)}$ | The weight of particle $i$, at time $t$ |
| $N_t$ | The total number of particles at time $t$ |
| $\delta_x(\cdot)$ | The delta-Dirac mass located in $x$ |
| $\mathcal{N}(\cdot : a, b)$ | Gaussian density with mean $a$ and covariance $b$, |

| $K_h(\cdot)$ | A kernel function with bandwidth $h$ |
|---|---|

Nonlinear filtering is a class of signal processing that widely exists in engineering, and is therefore a very broad research topic. The solution of the continuous time filtering problem can be represented as a ratio of two expectations of certain functions of the signal process. However, in practice, only the values of the observation corresponding to a discrete time partition are available; the continuous-time dynamic system has to be converted into a discrete-time simulation model, e.g. discrete Markov System, by discretely sampling the outputs through discretization. This paper is concerned with the problem of discrete filtering, which can be described in the State Space Model (SSM) that consists of two equations:

$$x_t = g_t\left(t, x_{t-1}, u_{t-1}\right) \qquad \text{(state transition equation)} \tag{1}$$

$$y_t = h_t\left(t, x_t, v_t\right) \qquad \text{(observation equation)} \tag{2}$$

The filtering problem recursively solving the marginal posterior density $p(x_t|Y_t)$ can be determined by the *recursive Bayesian estimation*, which has two steps:

(1) Prediction

$$p\left(x_t|Y_{t-1}\right) = \int_{\Re} p\left(x_t|x_{t-1}\right) p\left(x_{t-1}|Y_{t-1}\right) dx_{t-1} \tag{3}$$

(2) Updating or correction

$$p\left(x_t|Y_t\right) = \frac{p\left(y_t|x_t\right) p\left(x_t|Y_{t-1}\right)}{\int_{\Re} p\left(y_t|x_t\right) p\left(x_t|Y_{t-1}\right) dx_t} \tag{4}$$

In (3) and (4), the integration of often unknown and maybe high-dimensional functions is required, which can be computationally very difficult. This makes analytic optimal solutions such as the Kalman filter intractable. One flexible solution is the Monte Carlo approach, as the topic of this paper, which uses random number generation to compute integrals. That is, the integral, expressed as an expectation of $f(x)$ over the density $p(x)$, is approximated by a number of random variables $x^{(1)}, x^{(2)}, \dots x^{(N)}$ (called samples or particles) that are drawn from the density $p(x)$ (if possible), then

$$\hat{f} = \int f(x) p(x) dx = \mathrm{E}_{p(x)} \left[ f(x) \right] \approx \frac{1}{N} \sum_{s=1}^{N} f\left( x^{(i)} \right), \ x^{(i)} \sim \tag{5}$$

This is an unbiased estimate and, provided the variance of $f(x)$ is finite, it has a variance which is proportional to $1/N$.

However, one limitation in applying Monte Carlo integration (5) in Bayesian inference (3) and (4) is that sampling directly from $p(x)$ is difficult, even impossible, if high density regions in $p(x)$ do not match up $f(x)$ with areas where it has a large magnitude. A convenient solution for this is the *Importance Sampling* (IS). Assuming the density $q(x)$ roughly approximates the density (of interest) $p(x)$, then

$$\hat{f} = \int f(x) \left( \frac{p(x)}{q(x)} \right) q(x) dx = \mathrm{E}_{q(x)} \left\{ f(x) \left( \frac{p(x)}{q(x)} \right) \right\} \tag{6}$$

This forms the basis of Monte Carlo IS which uses the weight sum of a set of samples from $q(x)$ to approximate (6):

$$\hat{f} \approx \frac{1}{S} \sum_{s=1}^{S} f\left( x^{(s)} \right) \left( \frac{p\left( x^{(s)} \right)}{q\left( x^{(s)} \right)} \right) \tag{7}$$

An alternative formulation of IS is to use

$$\hat{f} \approx \hat{I} = \sum_{i=1}^{N} w^{(i)} f\left( x^{(i)} \right) / \sum_{i=1}^{N} w^{(i)}, \quad w^{(i)} = \frac{p\left( x^{(i)} \right)}{q\left( x^{(i)} \right)} \tag{8}$$

with the variance given by

$$\mathrm{var}\left[ \hat{f} \right] = \left( w^{(i)} \right)^2 \left[ \int \frac{\left( f(x) p(x) \right)^2}{q(x)} dx - \mathrm{E}_p \left\{ f(x) \right\} \right] \tag{9}$$

where $x^{(i)}$ is drawn from the proposal density $q(x)$. The variance is minimized to zero if $p(x) = q(x)$ [20, 24]. There are many potential choices for $q(x)$ leading to various integration and optimization algorithms, as shown in the summary provided in [38]. In general, $q(x)$ should have a relatively heavy tail so that it is insensitive to the outliers.

In the importance sampling the estimator not only depends on the values of $p(x)$ but also on the entirely arbitrary choice of the proposal density $q(x)$. This results in heavy dependence on irrelevant information $q(x)$. This *sampling difficulty* seems inevitable as the density $p(x)$ of interest is generally always unknown; therefore, it is impossible to direct the sampling. For this reason, some advanced important sampling methods have been proposed, such as annealed importance sampling [25], Bayesian importance sampling [24], adaptive importance sampling [34], numerically accelerated importance sampling [26], and nonparametric importance sampling [22]. Alternatively, sampling strategies such as rejection sampling [27], block sampling [28], Markov Chain Monte Carlo (MCMC) sampling [37, 38] and factored sampling [4, 29] have also been used, in addition to ad-hoc strategies such as multiple stages of important sampling [35]. SMC samplers are specifically developed to sample sequentially from a sequence of probability distributions that allows the calculation of a weight update equation for different proposal kernel [38]. To avoid an overly broad discussion, the reader is referred to the given references for further details.

The importance sampling implemented under the recursive Bayesian interference is referred to as Sequential Importance Sampling (SIS), which is the basis of PF. Formally, the core idea of PFs is to capture the statistics of the state probability distribution by a set of random particles with associated weights, i.e.

$$p\left(x_t \middle| Y_t\right) \approx \sum_{i=1}^{N_t} w_t^{(i)} \delta_{x_t^{(i)}}\left(x_t\right) \tag{10}$$

$$w_t^{(i)} \propto w_{t-1}^{(i)} \frac{p\left(y_t \middle| x_t^{(i)}\right) p\left(x_t^{(i)} \middle| x_{t-1}^{(i)}\right)}{q\left(x_t^{(i)} \middle| x_{t-1}^{(i)}, y_t\right)} \tag{11}$$

The weights will be normalized to sum to one (for the one-state output case).

Classical Monte Carlo procedures entirely ignore the state values of particles in the state space when forming the estimate, see [24]. In the process of propagation, particles perform a step of Markov jump for prediction and then the approximation density need to be adjusted through re-weighting the particles. After a few iterations in the particle propagation process, the weight will concentrate on a few particles only and most particles will have negligible weight, resulting in namely *sample degeneracy,* see [30]. This is an inherent default of the SIS. To address sample degeneracy, the standard PF is commonly accompanied with the resampling procedure (referred to Sampling-Importance Resampling, SIR or Sequential Importance Sampling and Resampling, SISR) to force particles to areas of high likelihood by multiplying high weighted particles while abandoning low weighted particles. This, however, may cause another problem: *sample impoverishment,* which occurs when very few particles have significant weight while most other particles with small weight are abandoned during the resampling process. To alleviate this, it is possible to balance the trade-off by applying resampling only at deterministic steps; that is, to execute resampling only when the variance of the non-normalized weights is superior to a pre-specified threshold (which is taken as a signal of sample degeneracy). In the general single-target case, a simple estimation of the Effective Sample Size (ESS) criterion is given by

$$ESS = \left( \sum_{i=1}^{N_t} \left( w_t^{(i)} \right)^2 \right)^{-1} \text{ s.t. } \quad \sum_{i=1}^{N_t} w_t^{(i)} = 1 \qquad (12)$$

The *ESS* takes values between 1 and $N_t$ and the resampling is implemented only when it is below a threshold $N_T$, e.g. $N_T = N_t/2$, $N_t$ is the sample size at time $t$.

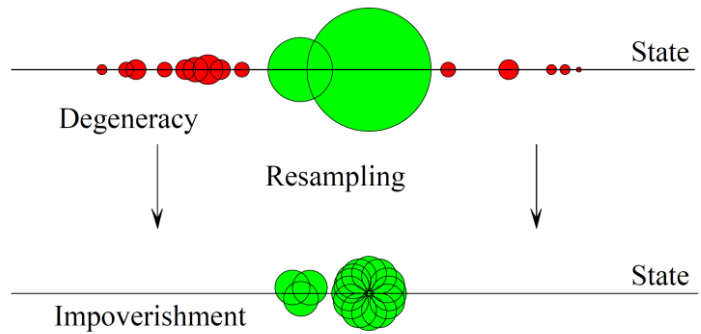

Figure 1. Sample degeneracy and impoverishment illustrated in 1-dimensional state space

The relationship between degeneracy and impoverishment can be depicted as shown in Figure 1 in which the size of the circles represents the weight of the particles; in the bottom row, connected circles share the same state after generic resampling. In the resampling process, only particles with significant weight (shown in green in Figure 1) are sampled, while other small weighted particles (shown in red) are abandoned. As shown therein, sample degeneracy is apparently the result of particles being distributed too widely, while sample impoverishment can be viewed as particles being over concentrated. The degeneracy converts to impoverishment as a direct result of resampling. If the resampling is unbiased, then the more serious the degeneracy, the more serious the impoverishment. In more general terms to remark on their relativity, sample degeneracy and impoverishment are a pair of contradictions that can be collectively described as a trade-off between the need of diversity and the need of focus in [32], as a problem in managing the spread of the particle set in the state space to balance conflicting requirements in [33], or in another perspective, as the computing resource converging prematurely either by weight or by state in [30].

Sample degeneracy and impoverishment are arguably a pair of fundamental difficulties of PFs, which manifests themselves as 'unsatisfactory' particle distribution. Quite intuitively, the solution to mitigate this problem is to optimize the distribution of particles either in advance or in hindsight. This forms the basis of our survey. In what follows, efforts devoted in this discipline are reviewed in different categorizes to illustrate the motivation, implementations and implications of PDO. Their interrelationships, pros, cons and high-dimensionality challenge are provided.

### 3. PARTICLE DISTRIBUTION OPTIMIZATION

To deal with sample degeneracy and impoverishment, there are mainly two kinds of information about the weighted particles that can be taken into account when one optimization operation is executed: weight and state. The PDO approaches presented in this paper are especially interested in the state of particles. Compared to distribution smoothing methods, most approaches act in a more intelligent manner and benefit from a variety of intelligent or heuristic techniques such as MCMC methods, Mean-shift algorithms, Artificial Intelligence (AI) algorithms e.g. Particle Swarm Optimization (PSO), Genetic Algorithm (GA) and Ant Colony Optimization (ACO) and Machine Learning (ML) approaches e.g., clustering, splitting and merging, etc. They form a coherent perspective to optimize the distribution of particles.

In general terms, the moving of particles can be '*blind*' (particle are moved without a specific direction) or '*sighted*' (particles are moved in a specific direction). In the latter case, new observations are used to direct the moving operation that can be referred to as *data-driven* methods, which is the main part of this study. In particular, various AI and ML optimizations for PF are specially reviewed. It is worth noting that, there is generally no standard implementation of each method; instead, one intelligent technique can be applied in many different ways within PF. In this case, too much is as bad as too little. The study emphasizes the similarity of all works in order to form the common PDO principle, and it will not cover all of the possible implementations. The primary PDO techniques that will be reviewed in this paper are categorized as shown in Table II. In each category, typical examples will be given as an explanation if possible. The basic idea will be briefly introduced first, and then the interrelationship and special achievements of different implementations are presented for further elaboration.

Table II. Categories of primary PDO approaches to review

| PDO tool | | Typical PDO property | Primary references |
|---|---|---|---|
| Kernel smoothing | | Blind | [41-45, 48] |
| Data-driven method | MCMC | Sighted | [16, 25, 33, 37, 40, 50-52] |
| | Mean-shift | Sighted | [33, 53-56] |
| AI algorithms (Evolution and population) | GA | Blind | [57-60, 71] |
| | PSO | Sighted | [63-66, 91] |
| | ACO | Sighted | [67-70] |
| ML techniques | Clustering | Blind | [53, 77-83] |
| | Merging/splitting | Blind | [36, 39, 76] |
| | Others | scatter search process [61, 72], support vector machines & support vector data description [29], etc. | |

| Hybrid PDO | Evolution + clustering [77], artificial immune system + PSO [92], kernel mean-shift + annealed PF [33], artificial neural networks + genetic algorithm [89]. |
|---|---|

### 3.1. Roughening, Kernel smoothing and regularization

Resampling, which is a type of re-selection and re-weighting of particles [84], is originally adopted to force particles to areas of high likelihood from low likelihood areas. However, as early as resampling was first proposed for PF in [42], its side effect (i.e. sample impoverishment) was apparently found to cause gaps between particles and accordingly a solution called *roughening* was proposed to smooth the posterior. Smoothness, here acting as the fitness, results in a continuous state probability distribution, which permits a better distribution diversity of the particles [15, 41].

The roughening procedure (the terms *jittering* [43], *diffusing* [72], *diversifying* [87] etc. are also used) basically adds an independent Gaussian jitter noise with zero mean and constant covariance, say $J_t$, to each resampled particle. Suppose that the original posterior density is denoted as $p(x_t|y_{0:t})$. Since the addition of two independent random variables corresponds to a convolution operation in the density domain, the approximate posterior density $\tilde{p}(x_t|y_{0:t})$ obtained after the roughening process can be factored as [39]

$$\tilde{\ }_t \approx \left\langle \mathcal{N}\left(x_t:0,J_t\right), p\left(x_t|y_{0:t}\right)\right\rangle \tag{13}$$

where $<\cdot>$ denotes the convolution operation, defined as $\left\langle f,g\right\rangle(x)=\int f(u)g(x-u)du$. Simply, if $p(x_t|y_{0:t})$ is Gaussian with mean $\tilde{x}_t$ and covariance $P_t$, then

$$\tilde{\ }_t \qquad \mathcal{N} \qquad \tilde{\ } \tag{14}$$

In (14) we can see that, the roughening strategy may be implemented more directly by increasing the simulation noise of the dynamic propagation of particles, called direct roughening [48]. Furthermore, in target tracking cases, roughening may be employed only in selected steps, on partial particles and in partial dimensions [84]. It is also suggested to employ the observation noise to scale the jittering strength so that jittering would not blur the position estimates out of the range of the observation.

With regard to scattering the multiple copies of the same particles, roughening is very equivalent to using a Gaussian kernel to smooth the posterior density. The basic principle of kernel smoothing is that local averaging or smoothing is performed with respect to a kernel function. To implement kernel smoothing mathematically, each particle is convolved with a diffusion kernel. The distribution is given as follows

$$p\left(x_t | y_{0:t}\right) \approx \sum_{i=1}^{N_t} w_t^{(i)} K_h\left(x_t - x_t^{(i)}\right) \qquad (15)$$

with the rescaled Kernel density

$$K_h\left(x\right) = \frac{1}{h^d} K\left(\frac{x}{h}\right) \qquad (16)$$

where the bandwidth $h > 0$, $d$ is the dimensionality. Under mild conditions ($h$ must decrease with increasing $N_t$) the kernel estimate converges in probability to the true density. The kernel and bandwidth are chosen so as to minimize the mean intergraded error or the mean integrated square error between the posterior distribution and the corresponding regularized weighted empirical measure.

Based on the Kernel method, the so-called *regularization* (of the empirical distribution associated to the particles) technique calculates a continuous analytical expression for the particle probability distribution [14]. There are two different approximations called pre-regularized PF (pre-RPF) and post-regularized PF (post-RPF), depending on whether the regularization step is taken before or after the correction step, see [44]. While the optimal kernel is intuitively appealing, and also satisfies an optimality criterion of some sort, it should be noted that it is possible to sample directly from such a kernel and to evaluate the weight integral analytically only in specific classes of model [15]. The selection of smoothing parameters (e.g. kernel bandwidth or roughening variance) is customized to specific problems and seems easier than it is. For a special review of the kernel based PDF estimation algorithms and their respective performance, the reader is referred to [45].

### 3.2. Data-driven methods

Kernel smoothing is a straightforward albeit 'blind' way to rejuvenate the diversity of particles to fight the sample degeneracy and impoverishment, in which no new information is employed. It is more reasonable to adjust the distribution of particles in a data-driven/'sighted' manner, e.g. move the particle to a better position by using the newest observations, namely new-observation-driven methods as discussed below.

The aim of new-observation-driven PDO methods is to use the newest observation, in which the particles tend to cluster in regions where the conditional posterior distribution for the current state has a high probability mass. There are two well-known data-driven approaches to building better proposal density. One is the auxiliary variable method [46] which augments the existing "good" particles in the sense that the predictive likelihoods are large for the "good" particles, which is quite similar to the *prior editing* proposed in [42]. The second approach uses the look-ahead strategies [47] or local perturbed sampling [49] to construct efficient proposal distributions. Both incorporate the information of the state dynamics and the current observation to combat the blindness of SIS. Much of this content is well documented in literature and will not be repeated here.

We will now consider some of the more profound variants of the principles exposed so far for sampling from the desired distribution and/or improving the diversity of particles. They are *Markov chain transition, mean-shift* and some artificial intelligence algorithms.

***MCMC***: MCMC methods including random walk Monte Carlo are a class of algorithms for sampling from probability distributions based on constructing a Markov Chain that has the desired distribution as its equilibrium distribution. Relying upon Markov kernels with appropriate invariant distributions, MCMC will generate collections of correlated samples. To cope with the *sampling difficulty,* since we wish particles to be drawn from $p(x_{1:t}|y_{1:t})$, it seems reasonable to design Markov chain transition kernels, having $p(x_{1:t}|y_{1:t})$ as the stationary distribution. As with other Monte Carlo methods, the empirical average taken over the samples is used to estimate the expectation of interest. Unlike sequential importance sampling, the samples from MCMC are exact (drawn from the desired distribution) and outstandingly, are free of sample degeneracy and impoverishment, although the MCMC suffers from other disadvantages. One of them is that, MCMC methods cannot be used in an online sequential Bayesian estimation context [38].

As with IS, where a good importance function encourages more samples to be drawn from high probability regions, MCMC random walks spend more time in regions of the parameter space with high probabilities, producing more samples from those areas. There are many well-known ways to achieve this, including the Metropolis-Hastings (MH) methods and Gibbs sampler [18]. Denoting invariant distribution $p(x)$ and proposal distribution $q(x^*|x^{(i)})$, which involves sampling a candidate value $x^*$ given the current value $x^{(i)}$, the Markov chain then moves towards $x^*$ with acceptance probability

$$A\left(x^* \leftarrow x^{(i)}\right) = \min\left\{1, \frac{p\left(x^*\right)q\left(x^{(i)}|x^*\right)}{p\left(x^{(i)}\right)q\left(x^*|x^{(i)}\right)}\right\} \tag{17}$$

otherwise it remains at $x^{(i)}$. In any case, the algorithm will tend to favor samples that increase the likelihood ratio, however its stochastic nature allows it to sometimes accept values that decrease the likelihood ratio, allowing it to escape local minima. In mathematics, the transition kernel for the MH algorithm is

$$K_{\text{MH}}\left(x^{(i+1)}|x^{(i)}\right) = q\left(x^{(i+1)}|x^{(i)}\right)A\left(x^{(i)} \leftarrow x^{(i+1)}\right) + \delta_{x^{(i)}}\left(x^{(i+1)}\right)r\left(x^{(i)}\right) \tag{18}$$

where $r(x^{(i)})$ is the term associated with rejection

$$r\left(x^{(i)}\right) = \int q\left(x^*|x^{(i)}\right)\left(1 - A\left(x^{(i)} \leftarrow x^*\right)\right)dx^* \tag{19}$$

However, there are situations where a very large number of MCMC iterations would be required in order to reach the target distribution, especially when the likelihood for the new data point is far from the points sampled from the importance distribution $q(x_t|x_{1:t-1})$. To overcome this, a series of smaller transitions is replaced by a single large transition in [50]; the Monte Carlo variation of the importance weights is also reduced there. Furthermore, the simulated annealing [25] can be used in MCMC for handling isolated modes and finding the maximum of a complex function with multiple peaks. Simulated annealing is very closely related to Metropolis sampling (but does not require designing the proposal distribution in MH sampling), differing only in that the probability $A(x^* \leftarrow x^{(i)})$ of a move is given by

$$A\left(x^* \leftarrow x^{(i)}\right) = \min\left\{1, \left(\frac{p\left(x^*\right)}{p\left(x^{(i)}\right)}\right)^{1/T(t)}\right\} \tag{20}$$

where the function $T(t)$ is called the cooling schedule (setting $T = 1$ recovers Metropolis sampling). In fact, the simulated annealing strategy itself can be employed in PF, such as in [33, 56] which will be described later in detail. It is generally difficult to assess when the Markov chain, even the simulated annealing, has reached its stationary regime and on the contrary it can become easily trapped in local modes.

In spite of the successful employment of the MCMC sampling to replace the important sampling [37, 38, 50], there are other areas where MCMC could benefit SMC, especially to rejuvenate the particle diversity ([51] for example) and in turn SMC can also benefit MCMC, typically such as particle MCMC [52]. For example, the Resample-Move algorithm that adds a MCMC move step after the resampling step of the SMC algorithm forms a principled way to jitter the particle locations, and thus reduce impoverishment [37]. In addition to the Resample-Move method, block sampling is proposed; it aims to sample only component $x_t$ at time $t$ in regions of high probability mass (while the previously-sampled values of the components $x_{t-L+1:t-1}$ sampled are simply discarded), and then use MCMC moves to rejuvenate $x_{t-L+1:t}$ after a resampling step [16, 28], where $L$ is the length of the lag. Both resample-move procedure and block sampling are often taken as correctly weighted Monte Carlo updating schemes.

The MCMC transition which is naturally parallel processing can be employed to execute resampling which is the primary obstacle for parallelization of PF [40]. For the parallelization of the resampling and PF, the reader is referred to [84] for a comprehensive review. We believe there will be more potential hybrid of MCMC and SMC for further benefits of both. Before proceeding to the AI and ML categories, the following section will review another class of data-driven method for PDO, based on the Mean-shift algorithm.

**Mean-shift**: Mean-shift is a gradient based iterative non-parameter optimization method for locating the maxima of a density function given discrete data sampled from that function. Given an initial estimate $x$ and a specified kernel function $K(\cdot)$ with bandwidth $h$, the weighted mean of the density is

$$m(x) = \frac{\sum_{x^{(i)} \in N(x)} K\left(\frac{x^{(i)} - x}{h}\right) x^{(i)}}{\sum_{x^{(i)} \in N(x)} K\left(\frac{x^{(i)} - x}{h}\right)} \tag{21}$$

where $N(x)$ is the neighborhood of $x$, a set of points for which $K(x) > 0$. In particular, $m(x) - x$ is called mean shift. The mean-shift algorithm recursively set $m(x)$ as $x$, and repeats the estimation until $m(x)$ converges to $x$. In this way, kernel mean-shift hill climbs towards the target, minimizing the distance between target and model candidates.

Particles are redistributed (termed as *herded* in [53], *moved* in [54] and *derived* in [55]) to their local mode of the posterior density (or observation) by similar mean-shift analysis in hybrid PFs [53, 54, 55]. For example, the kernel PF (KPF) [54] is similar to RPF in the sense that a kernel density estimate [44] is used to approximate the posterior PDF. However, unlike RPF, which uses samples from the kernel density estimate to replace the original particles, KPF estimates the gradient of the kernel density and moves particles toward the modes of the posterior by the mean-shift algorithm, leading to a more effective allocation of particles.

To summarize, the Kernel mean-shift can be viewed as an attempt to cluster spread particles with Kernel mean-shift; however, if PF tends towards an incorrect local maximum the mean-shift step will accelerate the process. To mitigate this, the kernel mean-Shift algorithm can be combined with the Annealed PF rather than with the basic PF as in [33]. The annealed PF [56] uses annealing to smooth out the evaluation function, making the global maximum clearer and allowing particles to spread further by increasing the process noise (inspired by the *roughening* approach). This will not be caught on local clutter since the mean-shift component could pull particles back towards the true target. This meticulous optimization of the distribution of particles shows, to a great extent, a type of 'intelligence', for which more direct solutions are presented in the following subsection.

### 3.3. AI Optimization: Evolution and Population

As a natural combination of artificial intelligence and single processing, the *Evolution* and *Population* optimization strategies rooted in AI algorithms may be employed for PDO; this will be referred to as AI PDO in this paper. The evolutionary heuristics and population-based search solves complex optimization problems by maintaining a population of candidate solutions, and are feasible for obtaining efficient PDO. Ever since earlier attempts to use the genetic algorithm filter were introduced, there have been many efforts devoted to this task [57], the Annealed PF [56] (also the annealed importance sampling [25]) and the LS-N-IPS [49]. Recently, these emerging AI-PF hybrid approaches have been extensively studied. In the following section, we review some representative studies to consider their common characteristics in order to develop a profound understanding of AI PDO. The AI algorithms that will be reviewed include GA, PSO and ACO. We will put emphasis on how these artificial intelligent algorithms work to optimize the distribution of particles to combat sample degeneracy and impoverishment.

***GA***: GA is governed by the *Schema Theorem*, which was originally derived from binary string representation of the genes of a chromosome within an individual. The Schema Theorem can be expressed as follows, see also [60].

$$m\left(\epsilon,t+1\right)\geq m\left(\epsilon,t\right)\frac{f\left(\epsilon\right)}{\bar{f}}\left(1-p_c\frac{\delta\left(\epsilon\right)}{L-1}\right)\left(1-p_m\right)^{o\left(\epsilon\right)} \tag{22}$$

where $m(\epsilon,t)$ is the number of schema $\epsilon$ at generation $t$, $f(\epsilon)$ is the average fitness of chromosomes having the same schema, $\bar{f}$ is the average fitness of the whole population, $p_c$ is the *Crossover* probability, $\delta(\epsilon)$ is the length of a schema, $L$ is the chromosome length, $p_m$ is the *Mutation* probability and $o(\epsilon)$ is the order of a schema.

The population of the GAs evolves in a competition for survival by different genetic operations including *Selection*, *Crossover*, and *Mutation*. GA is a Monte Carlo method. In accordance with the fitness values, the individuals are selected to undergo Crossover and Mutation and search for an optimal solution. Crossover pairs two individuals and mates them, and Mutation randomly alters the selected individual. Genetic operators are used in PF, either separately or in combination, to optimize the position of the particles by the genetic operations; they are used to deal with the situation in which most particles have collapsed at a single (i.e. sample impoverishment) point. These have been partly achieved since [57], and improved in different ways by [60], [71] and [58]. For an intuitive understanding, one evolutionary PF depicted in pseudo codes is given in [58] where a new form of the importance weight is derived. To further avoid premature concentration of the particles, the search region of particles can be enlarged within GA [59].

*PSO*: In the basic PSO algorithm, a set of particles are generated randomly, and their positions (states) are iteratively updated according to their own experience and the experience of the swarm (or neighboring particles). The particles are updated according to the following equations:

$$v_t = q v_{t-1} + \phi_1 \left( x_{ibest} - x_{t-1} \right) + \phi_2 \left( x_{gbest} - x_{t-1} \right) \tag{23}$$

$$x_t = x_{t-1} + v_t \tag{24}$$

where, $i$ is the current iteration step, $v_t$ is the flying speed of particles, $x_{ibest}$ is the particle's location at which the best fitness has been achieved, $x_{gbest}$ the population global location (or local neighborhood position $x_{nbest}$, in a neighborhood version of the algorithm) at which the best fitness so far has been achieved and $\emptyset_1$, $\emptyset_2$, $w$ are weighting factors. As a rule of thumb, the two random control factors $\emptyset_1$, $\emptyset_2$ in equation (23) are typically drawn from the uniform distribution $U(0, 2.05)$. A large inertia weight $q$ facilitates a global search while a small inertia weight facilitates a local search. As a result, the following linearly decreasing weighting function is usually utilized in (23)

$$q = q_{max} - \left( q_{max} - q_{min} \right) \times \frac{i}{I} \tag{25}$$

where, $q_{max}$ is the initial weight, $q_{min}$ is the final weight, and $I$ is maximum iteration number.

By exploring the likelihood distribution of the recent observations, the PSO particle flying strategy can help PF to obtain samples with high likelihoods [63, 64, 65]. The PSO algorithm distributes the particles in high likelihood area, regardless of the weight of particles in case of the dynamic model is unavailable. In addition, two different base points are used to distribute particles in order to achieve diversity and convergence in [63]. It is even shown that in a Bayesian inference view the sequential PSO framework is a swarm-intelligence guided multilayer importance sampling strategy [65]. However, applying PSO directly to PF involves two problems that have to be dealt with. One is the loss of particle diversity after the PSO procedure. To mitigate this, particles can be redistributed to increase the diversity after PSO [65]. The second problem is that a single swarm might not be enough due to the variation of the maximum likelihood point. To handle this, a *multi-swarm* mechanism maintaining multiple trajectories for a possible target position has been added to the generic PSO algorithm for robust tracking in [66].

Unlike the evolutionary programming and evolutionary strategies in GA, PSO does not implement the principle of the survival of the fittest, as there is no selection and crossover operation. To enhance this, the particle flying strategy combined with the mutation operation is proposed in [91] for PDO.

*ACO*: The ACO meta-heuristic that is initially proposed for solving combinatorial optimization problems (COPs) can benefit PF for PDO as well, especially after it is defined in the continuous domains ($ACO_{\mathbb{R}}$) [69]. The central component of ACO algorithms is the pheromone model, which is used to probabilistically sample the search space. ACO attempts to solve the problem by iterating two steps:

Step 1: A number of artificial ants build solutions to the problem by sampling a PDF which is derived from the pheromone information. In the basic Ant System (AS), the *i*th ant moves from state $x^{(i)}$ to state $x^*$ with probability

$$A\left(x^* \leftarrow x^{(i)}\right) = \frac{\tau^\alpha\left(x^* \leftarrow x^{(i)}\right)\eta^\beta\left(x^* \leftarrow x^{(i)}\right)}{\sum \tau^\alpha\left(x^* \leftarrow x^{(i)}\right)\eta^\beta\left(x^* \leftarrow x^{(i)}\right)} \tag{26}$$

where $\tau^{\alpha}(x^* \leftarrow x^{(i)})$ and $\eta^{\beta}(x^* \leftarrow x^{(i)})$ are, respectively, the amount of pheromone deposited and the desirability (heuristic value) associated with the state transition from state $x^{(i)}$ to state $x^*$ while $\alpha, \beta$ are positive real parameters whose values determine the relative importance of pheromone versus heuristic information.

Step 2: The solutions are used to modify the pheromone such that the probability to construct high quality solutions is increased. This is achieved by increasing the pheromone levels associated with chosen good solution $s_{ch}$ by a certain value $\Delta\tau$, and by decreasing all the pheromone values through pheromone evaporation

$$\tau^{\alpha}\left(x^* \leftarrow x^{(i)}\right) =$$
$$\begin{cases} (1-\rho)\tau^{\alpha}\left(x^* \leftarrow x^{(i)}\right) + \rho\Delta\tau & \text{if } \tau^{\alpha}\left(x^* \leftarrow x^{(i)}\right) \in s_{ch} \\ (1-\rho)\tau^{\alpha}\left(x^* \leftarrow x^{(i)}\right) & \text{otherwise} \end{cases} \quad (27)$$

where $\rho \in [0, 1]$ is the pheromone evaporation coefficient.

In addition to the above two steps, problem specific and/or centralized actions (Daemon actions) may be required [69]. ACO and ACO$_{\mathbb{R}}$ are incorporated into PF for moving particles to their local highest posterior density function in [70], which is to say, towards the region of the state space with the new observation [68]. The convergence result of ant stochastic decision based PF is presented in [67], in which each particle evolves in either of the two proposed ways to accommodate model variations. Particles are then selected (re-sampling) according to ant empiricism acquired from available observation. It is very interesting to notice that ACO PDO works in a data-driven manner that is very similar to the MCMC transition, mean-shift, and the PSO PDO as well. They all appear to have obvious *particle moving* character, differing only in implementation techniques and environments. However, there must be a balance point to optimize the particle distribution to avoid over-moving. This is reflected in the proper parameter setting evolved, as verified in the PSO embedded PF [66].

It should be noted that, very less evidence is available to demonstrate the convergence or even the optimality property of these AI PDO strategies reviewed so far within PF. This is because rigorous convergence analysis is often infeasible for heuristic and intelligent algorithms. Even so, one needs to be careful when applying intelligent techniques in PF otherwise the results may the exact opposite of what is desired. For this, reasonable parameter settings are critical to achieve the maximum benefit with the least side effects. This has been noticed in [69] in which the pheromone evaporation was applied to avoid too rapid a convergence of the GA algorithm (here the convergence is loosely defined). Further extension of evolution algorithms, called coevolution based on the interaction between species [73], is an option to preserve diversity within the population of evolutionary algorithms.

### 3.4.    ML Optimization: clustering, merging and splitting

The diversity of particles and the estimated uncertainty of PFs are essentially manifested in the spatial distribution (density) of particles, which is therefore worth considering for adjusting the sample size and maintaining the diversity of particles. A general idea for setting an appropriate sample size is to choose a small number of particles (i.e. samples) if the density is focused on a small part of the state space, otherwise a large number of particles should be chosen, see [75, 90]. Particles distributed in the same partition of the state space are considered to provide the same contribution to the diversity of particles [30], and more specifically, the Euclidean distance is used as a measure of the diversity of particles [61].

For example, when the state uncertainty is high, particles will be decentralized and distributed to a wider state space. When the state uncertainty is low, particles will be centralized and distributed to a small state space. Based on this understanding, the *KLD* (Kullback–Leibler Distance)-*sampling* approach [75] and KLD-resampling [90] determine the required number of particles $N_t$ so that the KLD between sample-based maximum likelihood estimate (MLE) and the true posterior does not exceed a pre-specified error bound threshold $\varepsilon$ by using the following equation

$$N_t = \frac{k-1}{2\varepsilon}\left(1 - \frac{2}{9(k-1)} + \sqrt{\frac{2}{9(k-1)}}z_{1-\delta}\right)^3 \tag{28}$$

where, $k$ is the number of grids with support, and $z_{1-\delta}$ is the upper quantile of the standard normal distribution. The equation shows that the number of particles is nearly proportional to the number of grid with support, which is based on the grid-partitioning of the state space. One case of grid partitioning in a 2-dimensional state space is shown in Figure 2.

In what follows, we consider a new class of PDO techniques that benefit much from techniques such as *clustering*, *merging* and *splitting* and that need to handle the dimensionality of the state space. Unlike the AI PDO techniques, these approaches may not be data-driven or heuristic, but are instead, apparently, ad-hoc as they are tailored to their particular applications e.g. robot localization (termed as Monte Carlo Localization, MCL) and target tracking, which are of relatively low dimensionality.

***Clustering*** is a natural logic analysis based statistic decision. By definition, cluster analysis or clustering is the assignment of a given set of data points into different groups, or clusters, based on some common properties of the points. The fact that spatially close particles represent a similar state raises the idea of clustering spatially close particles together to consider their common property. Clustering of particles is not only a means to reduce the sample size, but also a means to supervise the tracking convergence, to maintain the diversity of particles (with the potential ability to solve special problems such as the kidnapped-robot problem, which refers to a situation where the robot is carried to an arbitrary location and the tracking is completely lost) and to extract multi-estimates, etc.

One problem with MCL is that the plain bootstrap filter sometimes incorrectly converges to a unimodal distribution that is unable to maintain multimodal distributions. To solve this problem, spatially close particles are clustered together; each cluster is considered to be a hypothesis of the true state and is independently processed [79]. This allows for the solution of the kidnapped robot problem. While each cluster possesses a probability that represents the belief of the robot being at that location, the cluster with the highest probability would be used to determine the robot's location at that instant in time. It should be noted that both the filtering type and the clustering method are not specified and may be any advanced choice, such as the Uniform MCL [80], Sum-Of-Norms (SON) clustering [81].

The distribution of the particles provides an awareness about the degree of tracking convergence [78] which is helpful to know the progress of the localization in MCL. Based on this, the particle clustering technique is used to guarantee that the estimates are feasible at all times and positions [82]. The dynamical nature of clusters can be used to guarantee better coverage of the environment, allowing attention to be focused only where the probability to find the real robot is higher [53, 77]. In addition, in the multi-object tracking case based on the SMC-PHD (probability hypothesis density) filter [83], estimates are often extracted by peak searching of the particle distribution via clustering ($k$-means, Expected Maximum, etc.) albeit with its computational slowness and unreliability. A more sophisticated implementation of clustering allows more accurate and reliable estimate extraction, especially for the case of group/extended targets [23]. In contrast to employing clustering within PF, PF can in turn serve for clustering [74].

**Merging** and **Splitting**: Like the clustering technique, the *merging* and *splitting* (M&S) techniques are also based on the spatial similarity of particles to carry out PDO. They appear in variations based on different subjects: particles [36, 39] or particle clusters [76]. In the so-called particle M&S PF [36], different numbers of particles are used for prediction and updating separately to circumvent the contradiction between the estimation accuracy (the more particles, the better accuracy) and computational speed (the fewer particles, the faster speed). This works under the premise that weight updating is more computationally expensive than the prediction step, in which the computing speed will be highly improved when the number of particles for updating is reduced (by merging particles, see the left part of Fig. 2) while the prediction robustness and estimation accuracy can still be well maintained with a bigger prediction number of particles (by splitting particles, see the right part of Fig. 2). The merging and splitting in spirit execute a kind of threshold-based resampling [84] that can avoid the discarding of small-weighted particles. In addition, an appropriate smoothing e.g. roughening applied on the split particles, will be further helpful for the diversity of particles.

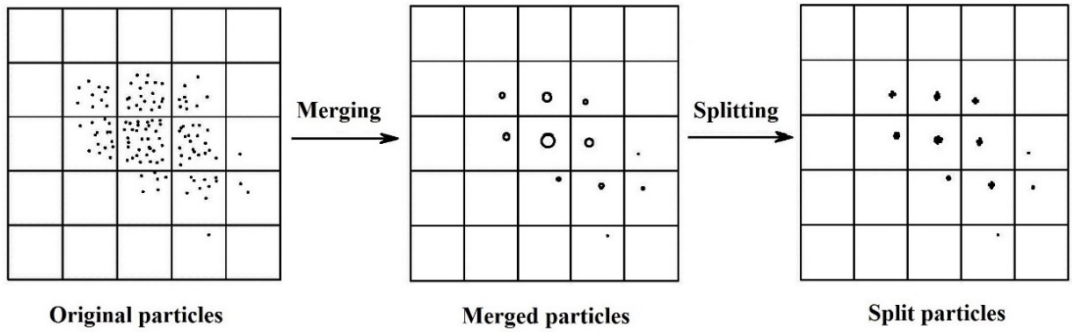

**Figure 2.** Particle merging and splitting illustrated in 2-dimensional state space

Furthermore, by means of *merging* and *splitting* the formed clusters in the Cluster PF, a dynamic clustered PF is proposed in [76]. Many other machine learning algorithms such as the scatter search process are available to execute PDO for PF; the implementation can also be quite flexible see [61, 72]. To improve the speed of the kernel density estimators in the aforementioned kernel smoothing strategy, machine learning approaches such as support vector machines (SVMs) and the support vector data description (SVDD) density estimation method [29] have been developed in PF. Despite tremendous effort, PDO based on machine learning still has much room to develop, not only for the benefit of better particle diversity (convergence and robustness) and faster speed, but also for improving estimation accuracy.

## 4. COMPUTATIONAL EFFICIENCY, HIGH DIMENSIONALITY AND BEYOND

We cannot expect to enumerate all the PDO effects not only because the literature is quite rich and will continually expand for some time to come, but also because it is hard to arrive at a rigorous definition of PDO that is simultaneously exclusive, exhaustive, separable and satisfying. To concentrate on a more coherent and intuitive understanding of PDO, less attention is placed on well-known work such as auxiliary variable methods [98], regularization methods [44, 99], decentralized and look-ahead PF [100]. The efforts we have reviewed so far allow a novel standpoint to improve PF that are particularly effective in dealing with sample degeneracy and impoverishment. In addition, the filtering reliability and convergence of PF might be improved accordingly. It is necessary to note that, the benefit of PDO techniques is generally problem-dependent and parameter-sensitive:

1) Each PDO technique may only work well on limited models, especially when it is initially designed for a specific problem, e.g. tracking lost (kidnapped robot) problem [79].

2) The benefits highly depend on the parameter setting. For the problem-specified PDO techniques, a very different performance may be achieved under different problem models or different parameter setting.

3) The benefits are not independent but instead highly interactively related to each other. e.g., to increase estimation accuracy is often accompanied with a decrease in computing speed. A more thorough discussion on the impact of computing speed to the estimation accuracy is given in the following subsection.

## 4.1 Hybrid PDO and Computational efficiency

Instead of utilizing a single PDO technique, there are some further hybrid approaches using two or more techniques to augment each other for hybrid optimization. For example, a method based on training artificial neural networks is introduced in [89] to implement the local search in the LS-N-IPS of [49]. Other hybrid PDO approaches include evolution along with clustering [77], artificial immune system with PSO [92], and kernel mean-shift algorithm with the annealed PF [33], etc., to name a few. There are in fact other ideas involved with PDO that may not seem so obvious.

However, it is necessary to note that approaches that are too complex may suffer from high computational burden that can heavily sacrifice the estimation quality in practice, although there does not appear to be anything wrong in simulations. This is because a complex filter design often accompanies a slowed-down sampling speed, which indicates a longer iteration period and heavier interval noise (e.g. the state transition noise). The increased noise can in turn sacrifice the performance of the filter. The good performance of complex filters, reported in many of the PDO strategies when using the same dynamic noise in the simulation, is highly suspicious. If the filtering speed is considered in practice, their improvements may not be obtained at all. This fact is overlooked in pure simulations where the simulation noise is constant regardless of the filtering speed. In fact, both theoretical and practical evidences show that choices that seem to be intuitively correct may lead to performances even worse than that of the plain bootstrap filter, see [62, 97]. This reminds us that thorough attention should be paid to the design and evaluation of a new filter, otherwise it may be overstated. For fair evaluation and comparison in simulations, the state transition noise should be simulated according to the sampling speed of each filter which is no easy task. This is elaborated in detail in [101]. For this reason, a traditional simulation comparison of different PDO approaches does not form part of our review.

## 4.2 High-dimensionality

It has been shown in [102-105] that, according to (8), the standard Monte Carlo error satisfies

$$\sup_{|f|\leq 1} \mathrm{E}|\pi_t(f) - \pi_t(\hat{f})| \leq \frac{C}{\sqrt{N}} \tag{29}$$

where $N$ is the number of particles, $\pi_t(\cdot)$ is the empirical measure function, the constant $C$ typically does not depend on time $t$ but must be exponential in the dimension of the state space of the underlying model. The exponential growth in the number of particles for increasing dimensions (known as the *cure-of-dimensionality*, see the evidence provided in [102]) is one of the biggest challenges for PDO, as well as one of the primary obstacles for the application of PF. It has been widely recognized that Monte Carlo methods may fail in large scale systems [88, 105], especially in geosciences [8, 106, 107] which could have one million or even more space–time dimensions.

The stress for PDO application to high dimensional systems is laid on space partitioning, indexing and searching of particles in the state space. To release this stress, advanced techniques and solutions can be roughly catalogued as follows: 1) reduce the performing dimensionality by functionally similar techniques such as Rao–Blackwellisation (RB) [108, 47, 93], decentralization [109], subspace hierarchy [110, 85], partitioned sampling [111], etc., 2) design heuristics procedures [72] suitable for high dimensional state space or curse-of-dimensionality-free operators [107], etc., 3) avoid the problem by employing more satisfied sampling of each particle to allow a small number of particles [106, 112] or by parallel computing [95]. We only provide brief introductions to each of these in the following paragraphs; for further details, readers are referred to the references provided.

The RB [108] approach partitions the state vector so that the Kalman filter is used for the part of the state space model that is taken as linear, while PF is used for the other part. For example, the state vector in the inertial navigation can have as many as 27 states, and here, the Kalman filter can be used for the 24 states, whereas PF is applied to the 3-D position state [17]. In order to remove the linear dependence using Kalman filter, the Decentralized PF (DPF) [109] splits the filtering problem into two nested sub-problems, and then handles the two nested sub-problems using PFs, which differs from RB in that the distribution of the second group of variables is also approximated by a conditional PF. Furthermore, the state space may be partitioned into more subspaces and run PF separately in each subspace [85]. Similarly, splitting the state space to reduce the important sampling dimension [88, 30], extracting hierarchical subspace to filter separately [110], partitioned sampling based on hierarchical search [111], and running SMC samplers in parallel in different regions of the state space with further possibility to interact with each other [113], are all similar and can be classified into the computationally efficient technique *partitioning* and *parallelization* to alleviate the burden of dimensionality in the high-dimensional search space as the terminologies suggest. It has been argued that it is often possible, at least in principle, to develop a local particle filtering algorithm whose approximation error is dimension-free [104].

On the other hand, efforts have been devoted to reinforce the use of each particle and thus to reduce the total number of particles required. A better proposal density using the so-called "nudging" method (future observations are employed in the proposal density to draw particles toward the observations) is explored, allowing the particles to know where the observations are in order to reduce the required number of particles [106]. Furthermore, Path Relinking (PR) and scatter search are evolutionary meta-heuristics that have been successfully applied to PF for high dimensional estimation problems, see [72]. The Smolyak operator underlying the sparse grids approach, which frees global approximation from the curse of dimensionality, is proposed for multivariate integration in [107]. These can afford more flexibility to PF for dealing with high dimensionality, especially in the case that only a small number of particles are allowed or preferred. Multiple 'Stochastic Meta-Descent' tracker are developed as 'smart particles' to track high-dimensional articulated structures with far fewer particles [114]. The box particle [115] occupies a small and controllable rectangular region having a non-zero volume in the state space which reduces the computational complexity and is suitable for solving high dimensional problems. Clearly, approaches that are effective for sample optimization in the state space of either low dimensionality or high dimensionality are all worth considering. As stated already, the resulting sampling speed of the filter should be taken into account.

## 5. Conclusions

This paper has reviewed a series of intelligent efforts that have been made on optimizing the distribution of particles in their propagation process in the particle filter. These efforts are particularly efficient for presenting or alleviating sample degeneracy and impoverishment. The survey emphasizes the similarity, interrelationships, pros and cons of these approaches rather than provide details on the variety of applications of each algorithm. An understanding of PDO was developed by considering all algorithms and techniques that share the same characteristics within a well-founded perspective, providing a systematic and coherent standpoint to study PF and allowing further improvements to be made. Some issues were not discussed including the rigorous reliability and convergence property of the PDO approach.

Finding more effective solutions for PDO in high dimensional problems remains, undoubtedly, an active and challenging topic in which there is still much more to do, especially in cases where only a small number of particles are allowed. For example, the newly appearing Cubature method [116] and quantum filtering [86] may have a potential benefit for PF. If high dimensionality is inevitable, the use of real-time techniques like parallel processing and ad-hoc techniques related to hardware improvement is suggested. Even, alternatives to the SIR such as MCMC are considerable.

We have noticed that most of advanced techniques used to enhance PF do not work for all cases but are problem-dependent and parameter-sensitive. In particular, the increased computational cost due to complex algorithm design may sacrifice performance quality in real life estimation, which is often overlooked in a pure simulation that uses constant parameters for all filters. As a result, the simulation outcome is not intended to represent the outcome of a real life situation. This deserves the attention of the users and designers of new discrete filters, not only PF. To overcome the gap between simulation and real life practice, a critical step involved is to setup the simulation model with respect to the computing speed of different filters. This is the key point to seamlessly connect simulation and reality and requires an urgent solution.

## ACKNOWLEDGMENT

This work was supported in part by the National Natural Science Foundation of China (Grant No.51075337; Grant No. 71271170) and the 111 Project (Grant No. B13044).